*Research Note*

# Potential-Based Shaping and Q-Value Initialization are Equivalent

**Eric Wiewiora**                                                           WIEWIORA@CS.UCSD.EDU
*Department of Computer Science and Engineering*
*University of California, San Diego*
*La Jolla, CA 92093 0114*

## Abstract

Shaping has proven to be a powerful but precarious means of improving reinforcement learning performance. Ng, Harada, and Russell (1999) proposed the *potential-based shaping* algorithm for adding shaping rewards in a way that guarantees the learner will learn optimal behavior.

In this note, we prove certain similarities between this shaping algorithm and the initialization step required for several reinforcement learning algorithms. More specifically, we prove that a reinforcement learner with initial Q-values based on the shaping algorithm's potential function make the same updates throughout learning as a learner receiving potential-based shaping rewards. We further prove that under a broad category of policies, the behavior of these two learners are indistinguishable. The comparison provides intuition on the theoretical properties of the shaping algorithm as well as a suggestion for a simpler method for capturing the algorithm's benefit. In addition, the equivalence raises previously unaddressed issues concerning the efficiency of learning with potential-based shaping.

## 1. Potential-Based Shaping

Shaping is a common technique for improving learning performance in reinforcement learning tasks. The idea of shaping is to provide the learner with supplemental rewards that encourage progress towards highly rewarding states in the environment. If these shaping rewards are applied arbitrarily, they run the risk of distracting the learner from the intended goals in the environment. In this case, the learner converges on a policy that is optimal in the presence of the shaping rewards, but suboptimal in terms of the original task.

Ng, Harada, and Russell (1999) proposed a method for adding shaping rewards in a way that guarantees the optimal policy maintains its optimality. They model a reinforcement learning task as a Markov Decision Process (MDP), where the learner tries to find a policy that maximizes discounted future reward (Sutton & Barto, 1998). They define a *potential function* $\Phi(\cdot)$ over the states. The shaping reward for transitioning from state $s$ to $s'$ is defined in terms of $\Phi$ as:

$$F(s, s') = \gamma\Phi(s') - \Phi(s),$$

where $\gamma$ is the MDP's discount rate. This shaping reward is added to the environmental reward for every state transition the learner experiences. The potential function can be viewed as defining a topography over the state space. The shaping reward for transitioning from one state to another is therefore the discounted change in this state potential. Potential-based shaping guarantees that no cycle through a sequence of states yields a net benefit from the shaping. In fact, under standard conditions Ng et al. prove that any policy that is optimal for an MDP augmented with a potential-based shaping reward will also be optimal for the unaugmented MDP.

## 2. New Results

Many reinforcement learning algorithms learn an optimal policy by maintaining Q-values. Q-values are estimates of the expected future reward of taking a given action in a given state. We show that the effects of potential-based shaping can be achieved by initializing a learner's Q-values with the state potential function. We prove this result only for the Q-learning algorithm, but the results extend to Sarsa and other TD algorithms as well.

We define two reinforcement learners, $L$ and $L'$, that will experience the same changes in Q-values throughout learning. Let the initial values of $L$'s Q-table be $Q(s, a) = Q_0(s, a)$. A potential-based shaping reward $F$ based on the potential function $\Phi$ will be applied during learning. The other learner, $L'$, will have a Q-table initialized to $Q'_0(s, a) = Q_0(s, a) + \Phi(s)$. This learner will not receive shaping rewards.

Let an experience be a 4-tuple $\langle s, a, r, s' \rangle$, representing a learner taking action $a$ in state $s$, transitioning to state $s'$ and receiving the reward $r$. Both learners' Q-values are updated based on an experience using the standard update rule for Q-learning. $Q(s, a)$ is updated with the potential-based shaping reward, while $Q'(s, a)$ is updated without the shaping reward:

$$Q(s, a) \;\leftarrow\; Q(s, a) + \alpha \underbrace{\big( r + F(s, s') + \gamma \max_{a'} Q(s', a') - Q(s, a) \big)}_{\delta Q(s, a)},$$

$$Q'(s, a) \;\leftarrow\; Q'(s, a) + \alpha \underbrace{\big( r + \gamma \max_{a'} Q'(s', a') - Q'(s, a) \big)}_{\delta Q'(s, a)}.$$

The above equations can be interpreted as updating the Q-values with an error term scaled by $\alpha$, the learning rate (assume the same $\alpha$ for the learners). We refer to the error terms as $\delta Q(s, a)$ and $\delta Q'(s, a)$. We also track the total change in $Q$ and $Q'$ during learning. The difference between the original and current values in $Q(s, a)$ and $Q'(s, a)$ are referred to as $\Delta Q(s, a)$ and $\Delta Q'(s, a)$, respectively. The Q-values for the learners can be represented as their initial values plus the change in those values that resulted from the updates:

$$
\begin{aligned}
Q(s, a) &= Q_0(s, a) + \Delta Q(s, a) \\
Q'(s, a) &= Q_0(s, a) + \Phi(s) + \Delta Q'(s, a).
\end{aligned}
$$

**Theorem 1** *Given the same sequence of experiences during learning, $\Delta Q(s, a)$ always equals $\Delta Q'(s, a)$.*

*Proof*: Proof by induction. The base case is when the Q-table entries for $s$ and $s'$ are still their initial values. The theorem holds for this case, because the entries in $\Delta Q$ and $\Delta Q'$ are both uniformly zero.

For the inductive case, assume that the entries $\Delta Q(s, a) = \Delta Q'(s, a)$ for all $s$ and $a$. We show that in response to experience $\langle s, a, r, s' \rangle$, the error terms $\delta Q(s, a)$ and $\delta Q'(s, a)$ are equal. First we examine the update performed on $Q(s, a)$ in the presence of the potential-based shaping reward:

$$
\begin{aligned}
\delta Q(s, a) &= r + F(s, s') + \gamma \max_{a'} Q(s', a') - Q(s, a) \\
&= r + \gamma \Phi(s') - \Phi(s) + \gamma \max_{a'} \left( Q_0(s', a') + \Delta Q(s', a') \right) - Q_0(s, a) - \Delta Q(s, a)
\end{aligned}
$$

Now we examine the update performed on $Q'$:

$$
\begin{aligned}
\delta Q'(s, a) &= r + \gamma \max_{a'} Q'(s', a') - Q'(s, a) \\
&= r + \gamma \max_{a'} \left( Q_0(s', a') + \Phi(s') + \Delta Q(s', a') \right) - Q_0(s, a) - \Phi(s) - \Delta Q(s, a) \\
&= r + \gamma \Phi(s') - \Phi(s) + \gamma \max_{a'} \left( Q_0(s', a') + \Delta Q(s', a') \right) - Q_0(s, a) - \Delta Q(s, a) \\
&= \delta Q(s, a)
\end{aligned}
$$

Both Q-tables are updated by the same value, and thus $\Delta Q(\cdot)$ and $\Delta Q'(\cdot)$ are equal. □

The implications of the proof can be appreciated when we consider how a learner chooses actions. Most policies are defined in terms of the learner's Q-values. We define an *advantage-based policy* as a policy that chooses an action in a given state with a probability that is determined by the differences of the Q-values for that state, not their absolute magnitude. Thus, if some constant is added to all the the Q-values, the probability distribution of the next action will not change.

**Theorem 2** *If $L$ and $L'$ have learned on the same sequence of experiences and use an advantage-based policy, they will have an identical probability distribution for their next action.*

*Proof:* Recall how the Q-values are defined:

$$
\begin{aligned}
Q(s, a) &= Q_0(s, a) + \Delta Q(s, a) \\
Q'(s, a) &= Q_0(s, a) + \Phi(s) + \Delta Q'(s, a)
\end{aligned}
$$

We proved that $\Delta Q(s, a)$ and $\Delta Q'(s, a)$ are equal if they have been updated with the same experiences. Therefore, the only difference between the two Q-tables is the addition of the state potentials in $Q'$. Because this addition is uniform across the actions in a given state, it does not affect the policy. □

It turns out that almost all policies used in reinforcement learning are advantage-based. The most important such policy is the greedy policy. The two most popular exploratory policies, $\epsilon$-greedy and Boltzmann soft-max, are also advantage-based. For any of these policies, there is no difference in learning between the initialization described above and potential-based shaping.

## 3. Shaping in Goal-Directed Tasks

It has been shown that the initial Q-values have a large influence on the efficiency of reinforcement learning for goal directed tasks (Koenig & Simmons, 1996). These problems are characterized by a state-space with some goal region. The agent's task is to find a policy that reaches this goal region as quickly as possible. Clearly an agent must find a goal state at least once during exploration before an optimal policy can be found. With Q-values initialized below their optimal value, an agent may require learning time exponential in the state and action space in order to find a goal state. However, in deterministic environments, an optimistic initialization of Q-values requires learning time that is polynomial in the state-action space before a goal is found. See Bertsekas and Tsitsiklis (1996) for further analysis of reinforcement learning algorithms with various initializations. Because potential-based shaping is equivalent to Q-value initialization, care must be taken in choosing a potential function that does not lead to poor learning performance.

## 4. Conclusion

We have shown that the effects of potential-based shaping can be captured by a particular initialization of Q-values for agents using Q-learning. These results extend to Sarsa and other TD methods. In addition, these results extend to the versions of these algorithms augmented by eligibility traces.

For a discrete-state environment, these results imply that one should simply initialize the learner's Q-values with the potential function rather than alter the learning algorithm to incorporate shaping rewards. In the case of continuous state-spaces, potential-based shaping may still offer some benefit. A continuous potential function over the state-space would be analogous to a continuous initialization of state values. Because potential-based shaping allows any function defined on the state space to be used as the potential function, the method may be beneficial to an agent with a restricted representation of state. A careful analysis of this case would be a fruitful avenue of future research.

## Acknowledgements

This research was supported by a grant from Matsushita Electric Industrial Co., Ltd.

## References

Bertsekas, D. P., & Tsitsiklis, J. T. (1996). *Neuro-dynamic Programming*. Athena Scientific.

Koenig, S., & Simmons, R. (1996). The effect of representation and knowledge on goal-directed exploration with reinforcement-learning algorithms. *Machine Learning*, *22*(1/3), 227 – 250.

Ng, A. Y., Harada, D., & Russell, S. (1999). Policy invariance under reward transformations: theory and application to reward shaping. In *Machine Learning, Proceedings of the Sixteenth International Conference*, pp. 278–287. Morgan Kaufmann.

Sutton, R. S., & Barto, A. G. (1998). *Reinforcement Learning: An Introduction*. The MIT Press.